\documentclass[conference]{IEEEtran}
\usepackage{graphicx}
\usepackage{xurl}
\usepackage{booktabs}
\usepackage{amsmath,amssymb}
\usepackage{tabularx}
\usepackage[table]{xcolor}
\usepackage[hidelinks]{hyperref}
\usepackage{dsfont}
\usepackage{float}
\usepackage{caption}
\usepackage[numbers]{natbib}
\IEEEoverridecommandlockouts

\title{Visual Word Sense Disambiguation with CLIP through Dual-Channel Text Prompting and Image Augmentations}

\author{
\IEEEauthorblockN{
Shamik Bhattacharya\IEEEauthorrefmark{1}\IEEEauthorrefmark{2},
Daniel Perkins\IEEEauthorrefmark{1}\IEEEauthorrefmark{2},
Yaren Dogan\IEEEauthorrefmark{3},
Vineeth Konjeti\IEEEauthorrefmark{3},
Sudarshan Srinivasan\IEEEauthorrefmark{4},
Edmon Begoli\IEEEauthorrefmark{3}\IEEEauthorrefmark{4}
}

\IEEEauthorblockA{
\IEEEauthorrefmark{1}Equal Contribution
}

\IEEEauthorblockA{
\IEEEauthorrefmark{2}The Bredesen Center for Interdisciplinary Research and Graduate Education, Knoxville, TN 37916
}

\IEEEauthorblockA{
\IEEEauthorrefmark{3}Department of Electrical Engineering \& Computer Science, University of Tennessee, Knoxville, TN 37916
}

\IEEEauthorblockA{
\IEEEauthorrefmark{4}Oak Ridge National Laboratory (ORNL), Oak Ridge, TN 37830
}

}

\begin{document}
\maketitle

\begin{abstract}
Ambiguity poses persistent challenges in natural language understanding for large language models (LLMs). To better understand how lexical ambiguity can be resolved through the visual domain, we develop an interpretable Visual Word Sense Disambiguation (VWSD) framework. The model leverages CLIP to project ambiguous language and candidate images into a shared multimodal space. We enrich textual embeddings using a dual-channel ensemble of semantic and photo-based prompts with WordNet synonyms, while image embeddings are refined through robust test-time augmentations. We then use cosine similarity to determine the image that best aligns with the ambiguous text. When evaluated on the SemEval-2023 VWSD dataset, enriching the embeddings raises the MRR from 0.7227 to 0.7590 and the Hit Rate from 0.5810 to 0.6220. Ablation studies reveal that dual-channel prompting provides strong, low-latency performance, whereas aggressive image augmentation yields only marginal gains. Additional experiments with WordNet definitions and multilingual prompt ensembles further suggest that noisy external signals tend to dilute semantic specificity, reinforcing the effectiveness of precise, CLIP-aligned prompts for visual word sense disambiguation.

\end{abstract}

\section{Introduction}
Ambiguity is the feature of a language that allows for multiple interpretations of the same expression, a property that has long challenged computer scientists and linguists alike. Ambiguity exists in many forms, including lexical, syntactic, semantic, and pragmatic \cite{li2024taxonomy}. Without sufficient context, even the most advanced large language models (LLMs) struggle to discern meaning in the intuitive way that humans do.

We focus on lexical ambiguity. It can be divided into two forms: polysemy, where a word has multiple related meanings, and homonymy, where a word has multiple unrelated meanings. Humans can distinguish between these meanings by leveraging the context clues around the word. For instance, in the phrase ``U.S. Army tank", the word ``tank" refers to a heavily armored military vehicle; whereas in ``water tank", it denotes a large container for storing liquid. 

LLMs first tokenize sentences and generate embeddings for each token, where words with similar meanings are mapped to similar locations in the vector space. Then, transformer-based LLMs, such as BERT \cite{devlin2019bertpretrainingdeepbidirectional}, leverage the embeddings of surrounding words to map the original token embeddings to contextual embeddings. When there is enough context, this captures the difference between meanings of the same word. However, when the context is limited in ambiguous sentences, these contextual embeddings are unlikely to represent the intended meaning.


Current research has focused on addressing this issue through Word Sense Disambiguation (WSD). WSD is the task of
associating a word in context with its intended
sense, generally from a pre-defined sense inventory \cite{7a59457361fa4188824d7d547bec7ff5}. Originally, WSD models generated embeddings for each provided definition of the given word and used the contextual embeddings to predict which definition is the most accurate via cosine similarity. This effectively characterizes the ambiguity of each target word. However, it is inherently limited by the surrounding words that the user provides as context.

This motivates the need to use other modes of information to resolve the ambiguity. In Visual Word Sense Disambiguation (VWSD), the goal is to find which image most accurately depicts the correct sense of the target word \cite{kwon2023vision}. This allows models to more accurately determine parse ambiguous statements when a visual input is present. For example, given the small phrase “internet router” with “router” as the target word and the ten candidate images, the model should identify which image best represents its intended sense.



To address these limitations, we propose a multi-modal approach to Visual Word Sense Disambiguation (VWSD) built on CLIP \cite{radford2021learningtransferablevisualmodels}. Rather than relying solely on raw cosine similarity between word and image embeddings, our method enhances semantic alignment by applying dual-channel textual prompts to better capture contextual nuance, incorporating lexicographical knowledge from WordNet, and using image augmentation to improve the robustness of visual embeddings. Together, these components yield more discriminative cross-modal representations for sense disambiguation.




\section{Previous Work}

Although many LLMs have been developed, limited research has been done on how to effectively resolve ambiguities in natural language processing (NLP). Many benchmark datasets have been introduced but there is inconsistent coverage of the different types of lexical ambiguity. Furthermore, no tool or technique can eliminate all ambiguities in natural language, especially in open-domain questions. 

\subsection{Word Sense Disambiguation (WSD)}

Initially, research primarily focused on leveraging context clues surrounding words for WSD. \citet{YADAV202185} comprehensively reviewed early methods describing a range of strategies including knowledge-based approaches, syntactic analysis, statistical methods, and heuristic-based approaches. Although these methods perform well in some cases, they require considerable effort and doesn't generalize well to unseen scenarios. Newer methods—including unsupervised learning, statistical normalization, and ensembles of LSTMs—have shown promising results for ambiguity detection \cite{abeysiriwardana2024surveylexicalambiguitydetection,info15090540,Ababor,10.1145/3477578} but are still limited by how much context is available in the surrounding words.

\subsection{Visual Word Sense Disambiguation (VWSD)}

\subsubsection{The Dataset}
\label{dataset}

Building on progress in WSD, recent research has shifted towards Visual Word Sense Disambiguation (VWSD). The SemEval-2023 competition introduced a dataset for VWSD \cite{7a59457361fa4188824d7d547bec7ff5} which allows for efficient model development and testing.

The dataset consists of tuples containing the target word, context phrase, the candidate images, and the gold image. The goal of this data set is to choose the image that best matches the meaning of the word in that context. Each target word comes with its own context phrase which specifies the ambiguity and 10 candidate images. The image that most represents the target word is also the gold image (correct choice).

\begin{figure}[ht]
    \centering
    \begin{minipage}[t]{0.28\textwidth}
        \centering
        \includegraphics[width=1\linewidth]{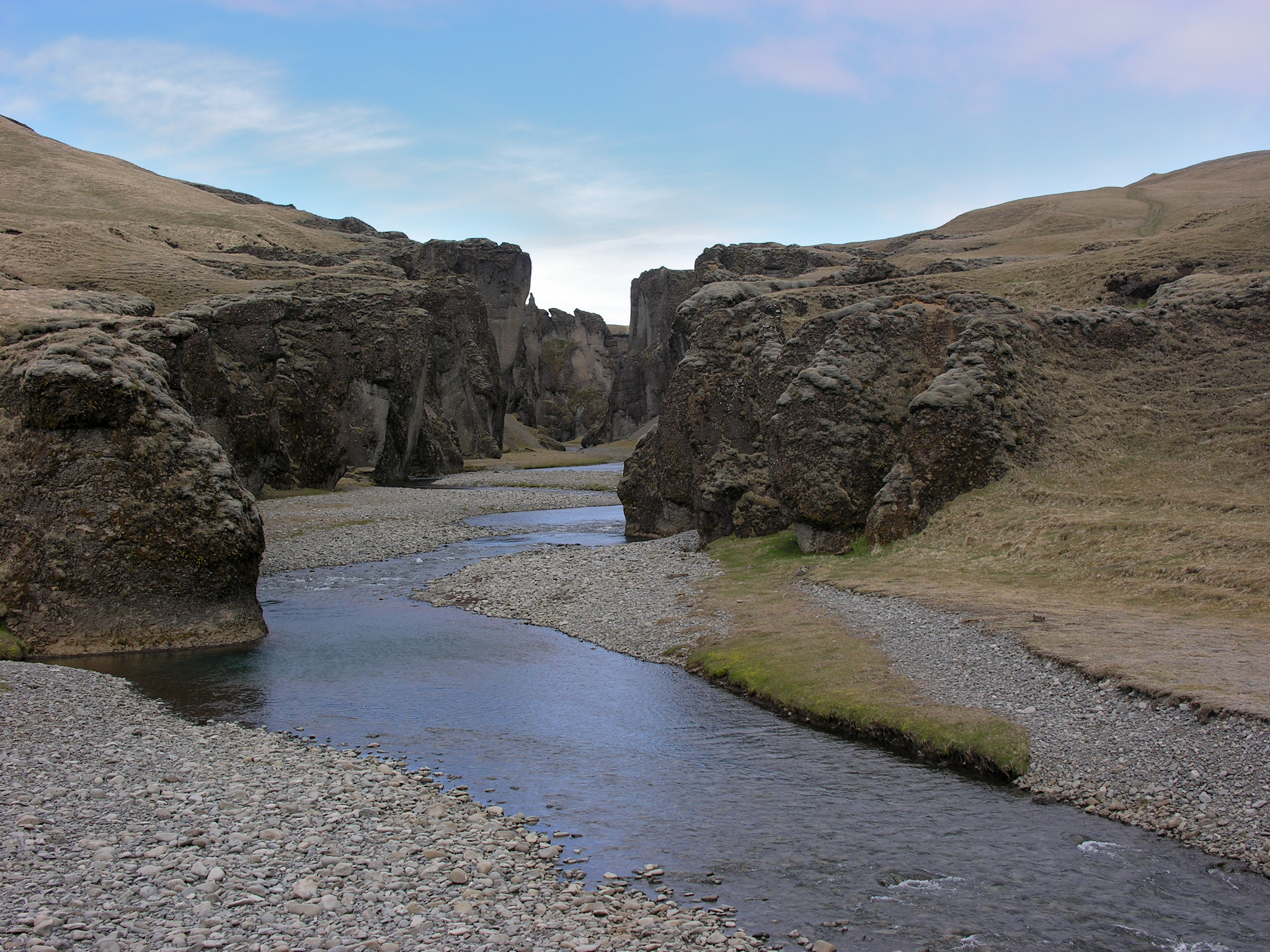}
    \end{minipage}
    \hfill
    \begin{minipage}[t]{0.16\textwidth}
        \centering
        \includegraphics[width=1\linewidth]{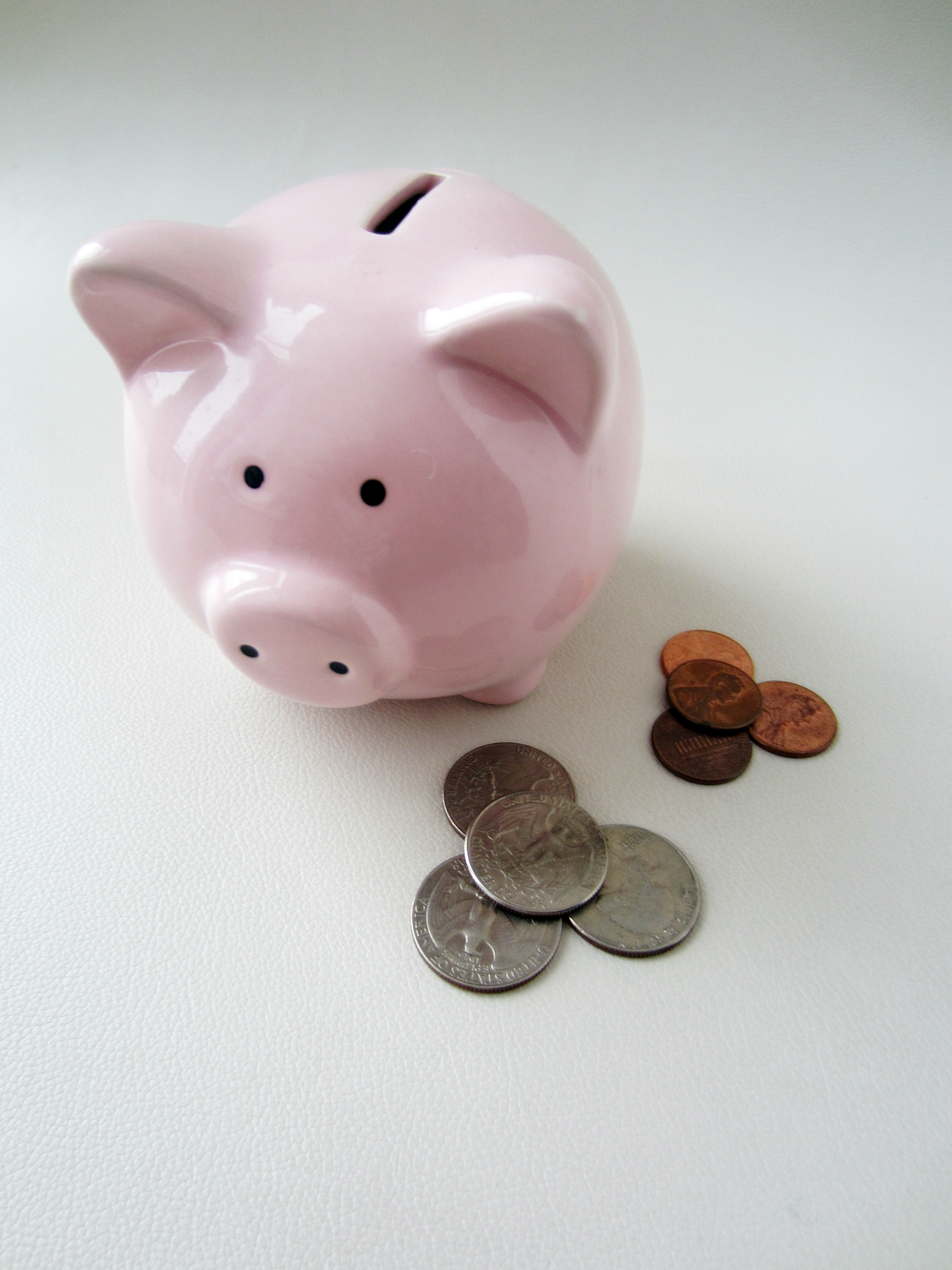}
    \end{minipage}
    \caption{Two images illustrating the ambiguity of the word ``bank'': one shows riverbank erosion, the other a piggy bank.}
    \label{fig:bank_ambiguity}
\end{figure}

For example, in the SemEval-2023 dataset, one element has the target word ``bank" with the context phrase ``bank erosion". Two of the 10 candidate images are provided in Figure \ref{fig:bank_ambiguity}. A correct model would choose the image on the left since it depicts erosion on a bank.

\subsubsection{Current Models}


Three top models from SemEval-2023’s Visual Word Sense Disambiguation task illustrate key advances:

\begin{itemize}
    \item The Samsung SRCB system \cite{zhang2023srcb} uses prompt-based context retrieval, large external sense inventories, and multimodal retrieval over external image collections to maximize performance, ranking first on the English track with an 84 percent hit rate.
    
    \item The ARPA hybrid model  \cite {papastavrou2024arpa} fused large language models with Swin Transformers and graph convolutional networks to capture fine-grained text–image relationships, achieving the best overall accuracy (84.7 percent) and MRR (88.7 percent) across multilingual datasets.
   
    \item A text-augmentation-based CLIP model \cite {kritharoula2023large} leveraged GPT-3-generated context and WordNet synonyms to improve zero-shot disambiguation, reaching high performance through enriched multimodal embeddings.
   
\end{itemize}

In contrast to these models, our goal is not to maximize raw accuracy, but to develop an interpretable framework that isolates the contributions of prompting, visual augmentations, and other strategies for enriching multimodal embeddings. Rather than relying on large external generators or complex multimodal fusion (e.g., GNNs and vision–LLM stacks), we use a dual-channel prompting scheme with WordNet synonyms and a controlled test-time augmentation pipeline. This design allows us to systematically evaluate how specific, low-cost design choices affect validation metrics under strict latency and complexity budgets.

\section{Approach}
\label{Approach}

\subsection{Vanilla CLIP-based Model}
\label{vanilla_model}

\subsubsection{Preprocessing}
\label{Preprocessing}

\begin{figure}[H]
    \begin{center}
        \includegraphics[width=0.85\linewidth]{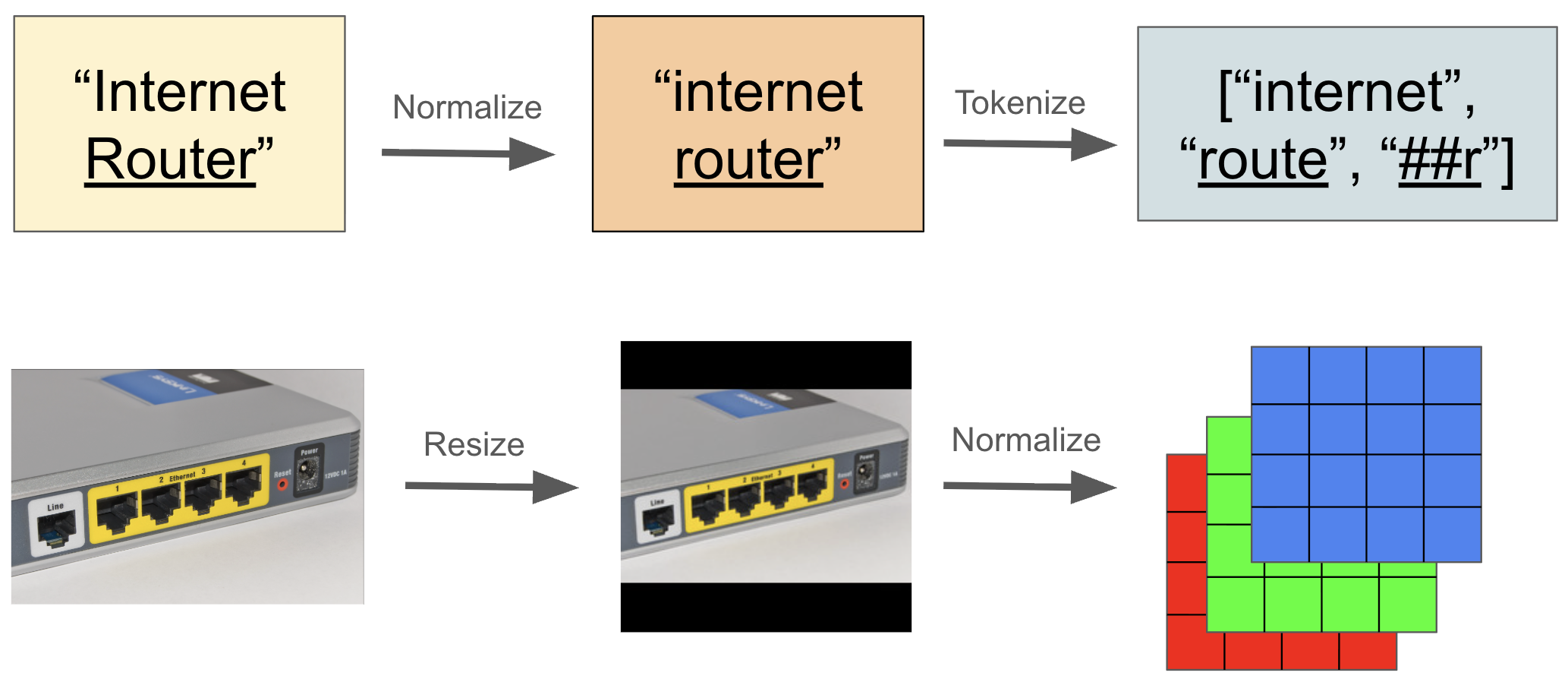}
        \caption{Normalization of the textual and visual input before they are passed into the vision language models. The sentence ``Internet Router'' (with the underlined target word ``router'') is normalized and tokenized. Additionally, the images are resized and normalized.}\label{fig:normalization}
    \end{center}
\end{figure}

We first preprocess all inputs to match the pre-training specifications of the CLIP architecture, as illustrated in Figure \ref{fig:normalization}. Text inputs are tokenized using CLIP's Byte-Pair Encoding (BPE) tokenizer with a maximum context length of 77 tokens. Candidate images are resized to a resolution of $224 \times 224$ and normalized using the standard mean and standard deviation values from the CLIP pre-training protocol.

\subsubsection{Contextual Embeddings}
\label{Contextual_Embeddings}

\begin{figure}[H]
    \begin{center}
        \includegraphics[width=1\linewidth]{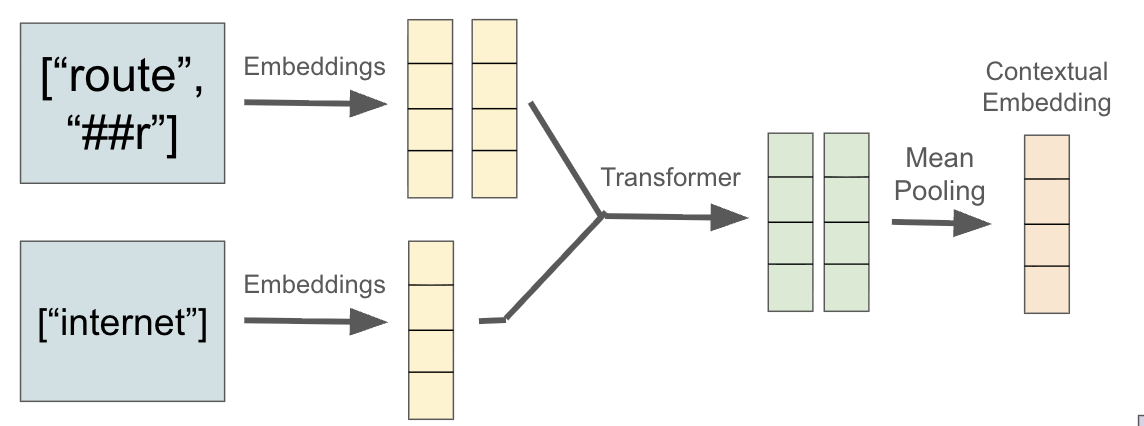}
        \caption{The pipeline for generating contextual embeddings. In this example, the phrase ``internet router" is tokenized. An initial embedding is created for the tokens of the target word (top) and the surrounding words (bottom). Each embedding is then passed into a transformer. Finally, the hidden states for the tokens of the target word are pooled together to create the contextual embedding. }\label{fig:ContextualEmbedding}
    \end{center}
\end{figure}

The next step is to derive contextual embeddings for the target word, as illustrated in Figure~\ref{fig:ContextualEmbedding}. 
The input text is first tokenized into subword units and mapped to initial token embeddings $h^{(0)}$ via the model’s 
learned embedding layer. The resulting sequence of embeddings is then processed by an encoder-based transformer, such as 
BERT~\cite{devlin2019bertpretrainingdeepbidirectional} or CLIP~\cite{radford2021learningtransferablevisualmodels}, which produces 
a hidden-state representation $h_i$ for each token in the sequence. The hidden state corresponding to the target word, denoted 
$h_t$, is taken as its contextual embedding. In cases where the target word is split into multiple subword tokens, the contextual 
embedding is computed as the mean of the corresponding hidden states.

Additionally, when no specific target word is defined (i.e., when the goal is to represent the overall meaning of the entire text), the contextual embedding is derived differently depending on the model architecture. When our model employs BERT-style encoders, the embedding is obtained by averaging the hidden states for all the tokens. When CLIP is used, since the hidden state for the end-of-text token serves as the global textual representation, it is used for the contextual embedding.

The embeddings produced by this pipeline capture the semantic meaning of the target word by integrating both its isolated meaning and the contextual information provided by the surrounding words in the sentence.

\subsubsection{Cosine Similarity} 
\label{cosine_sim}

The proposed architecture selects the image that best represents a target word given its surrounding context by jointly embedding textual and visual inputs into a shared multi-modal space. Textual inputs are encoded using a contextual sentence representation (Section \ref{Contextual_Embeddings}), while candidate images are embedded using a vision encoder. Both encoders are instantiated from the CLIP vision–language model, enabling direct similarity-based comparison between text and image embeddings. Semantic alignment is achieved through cosine similarity between the contextual text embedding and each candidate image embedding, with the highest-scoring image selected as the predicted sense.

In other words, given a target word $t$ and surrounding context word(s), we use CLIP to generate the contextual embedding $h_t$. And, given ten candidate images, $\{x_j\}_{j=1}^{10}$, we use CLIP to generate 10 visual embeddings $\{h_{x_j}\}_{j=1}^{10}$. The model selects the image the corresponds to the visual embedding with the highest cosine similarity to the contextual embedding:

\begin{equation}
    x_*=\text{argmax}_{j\in \{1,2,..,10\}}\frac{h_{t}\cdot h_{x_j}}{\|h_{t}\|_2 \|h_{x_j}\|_2}
    \label{original_equation}
\end{equation}

\begin{figure}[H]
    \begin{center}
        \includegraphics[width=1\linewidth]{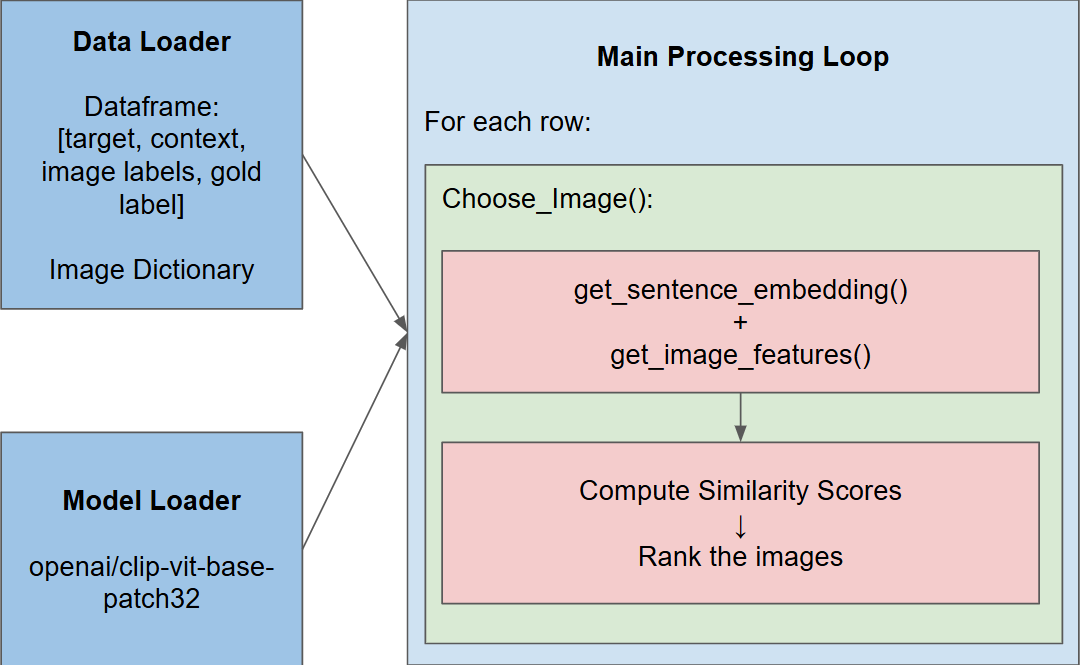}
        \caption{Overview of the vanilla CLIP-based model. Both text and image inputs are encoded into a shared multi-modal space, and cosine similarity determines which image best aligns with the contextual meaning of the target word.}\label{fig:CLIPapproach}
    \end{center}
\end{figure}

The implementation of this vanilla model is visualized in Figure \ref{fig:CLIPapproach} and the results are displayed in Table \ref{tab:clip_comparison}.

\subsection{The Proposed Architecture}
\label{proposed_architecture}

In our proposed model, we follow the framework explained in Section \ref{vanilla_model}. However, to more accurately resolve lexical ambiguity, before computing the cosine similarity, we enrich the original sentence $h_t$ and visual $h_{x_j}$ embeddings with prompt-based augmentations, WordNet synonyms, and image augmentations.

To reiterate, in the SemEval dataset, each point of the dataset includes a polysemous target word and a context word like shown in the examples below:

\begin{quote}
\textbf{Sentence:} \textit{internet router} \\
\textbf{Target:} \textit{router}

\vspace{0.5em}
\textbf{Sentence:} \textit{bank erosion} \\
\textbf{Target:} \textit{bank}

\end{quote}

\subsubsection{Prompting and Text-enrichment}

To enrich $h_t$, instead of directly using the given text in the dataset, we first generate a variety of semantically enriched prompts that highlight the ambiguous target word in other semantic contexts. Each prompt focuses on one interpretation, avoids cluttered definitions, stays close to CLIP's training distribution (short and photo-like), and reduces ambiguity between senses of the target word. Our prompts are split up into two channels:
\begin{itemize}
    \item \textbf{Semantic Prompts:} We employ multiple targeted prompts, $s_1,s_2,...,s_k$, to describe the target word in a narrow semantic angle. For example, the semantic prompts could include:
    \begin{itemize}
        \item ``bank related to erosion"
        \item ``the concept of bank in erosion"
        \item ``bank in the context of erosion"
    \end{itemize}
    \item \textbf{Photo Prompts:} To visually describe the word, we utilize prompts, $p_1,p_2,...,p_l$, that relate the ambiguous text to images. Some examples include:
    \begin{itemize}
        \item ``a photo of bank erosion"
        \item ``bank with erosion, natural scene"
        \item ``bank appearing in an erosion environment"
    \end{itemize}
\end{itemize}


Following the process explained in \ref{Contextual_Embeddings}, we generate embeddings for each of the semantic prompts $h_{s_1},h_{s_2},...,h_{s_k}$ and photo prompts $h_{p_1},h_{p_2},...,h_{p_l}$. For each channel, we compute the mean embeddings $\overline{h_s}=\frac{1}{k}\sum_{i=1}^kh_{s_i}$ and $\overline{h_p}=\frac{1}{l}\sum_{i=1}^lh_{p_i}$ and then normalize them to produce the semantic $h_s$ and photo $h_p$ embeddings. To generate the final textual embedding, these embeddings are then fused together with a weighted sum:
\begin{equation}
    \hat{h_t}=\beta_p h_p + \beta_s h_s
    \label{eq:dual_channel}
\end{equation}
Finally, $\hat{h_t}$ is L2-normalized to produce our final enriched textual embedding, $\tilde{h_t}$.

This technique acts as an ensemble inside CLIP's text encoder. The enriched text embedding models both visual concreteness and semantic relationship. The hyperparameter $\beta\in [0,1]$ determines the relative importance of each type of prompt, providing CLIP's text encoder with a robust way to fuse multiple linguistic angles.

\subsubsection{WordNet Synonoms}
\label{WordNet_Synonoms}

To further increase the variance of the input, we use WordNet to select one synonym of the target and another of the context word. For example, given the phrase ``bank erosion", we extract the WordNet synonyms ``edge" and ``deterioration". These new words are then used to generate more photo prompts, such as ``a photo of edge deterioration". These additional prompts are infused into the mean photo embedding $\overline{h_p}$, ultimately leading to a more robust enriched textual embedding $\tilde{h_t}$.


\subsubsection{Image Augmentations}
\label{augmen}


To enrich the image embeddings $h_{x_j}$, our model replaces single-view encoding with a comprehensive augmentation pipeline. Our augmentations combine both localized crops and global transformations so that the model can examine each candidate image from multiple spatial, photometric, and structural perspectives.

\begin{figure}[H]
\centering
\begin{tabular}{cc}
    \includegraphics[width=0.3\linewidth]{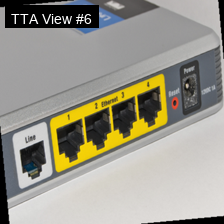} &
    \includegraphics[width=0.3\linewidth]{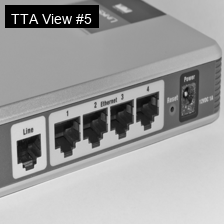} \\
    \small Random Rotation & \small Monochrome Coloring \\[1em]
\end{tabular}
\caption{Examples of the Test-Time Augmentation Strategy (monochrome coloring) and the Random Geometric Augmentation Strategy (random rotation).}
\label{fig:additional-tta}
\end{figure}

\begin{figure}[H]
\centering
\begin{tabular}{cc}
    \includegraphics[width=0.3\linewidth]{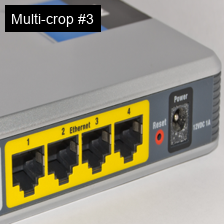} &
    \includegraphics[width=0.3\linewidth]{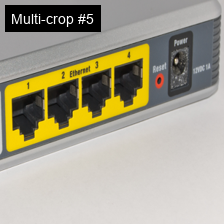} \\
    \small Zoom-in Image Centered & \small Zoom-in Image Raised \\[1em]
\end{tabular}
\caption{Examples of the Multi-Crop Extraction Strategy.}
\label{fig:multi-crop}
\end{figure}

\begin{figure}[H]
\centering
\begin{tabular}{cc}
    \includegraphics[width=0.3\linewidth]{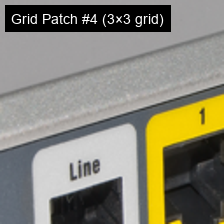} & 
    \includegraphics[width=.3\linewidth]{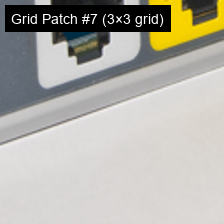} \\
    \small Left-middle Grid Patch & \small Lower-middle Grid Patch \\[1em]
\end{tabular}
\caption{Examples of the Grid Patches augmentation Strategy.}
\label{fig:grid-patch}
\end{figure}

\begin{figure}[H]
\centering
\begin{tabular}{cc}
    \includegraphics[width=0.3\linewidth]{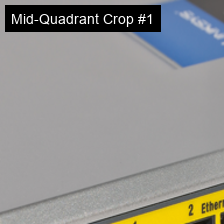} & 
    \includegraphics[width=.3\linewidth]{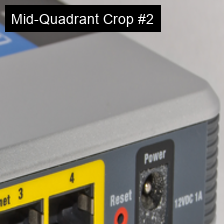} \\
    \small Left Mid-Quadrant Crop & \small Right Mid-Quadrant Crop \\[1em]
\end{tabular}
\caption{Examples of the Mid-Quadrant Crop augmentation Strategy.}
\label{fig:mid-quadrant-crop}
\end{figure}

We employ six image augmentation techniques, each designed to capture different visual variations for CLIP (see Appendix \ref{app:augmentations}). 
\begin{itemize}
\item \textbf{Test-Time Augmentation strategy:} This strategy includes many views of a single image. We have baseline views to ensure CLIP sees the unperturbed image, and using this baseline image, we mirror the image horizontally, center crop the image, zoom into the image, and remove any color from the image (Figure \ref{fig:deterministic-tta}). This makes the model robust to geometric and luminance variations. 
\item \textbf{Random Geometric Augmentations:} This strategy introduced perspective variations, almost mimicking a camera. These image variations include stochastic horizontal flip, small random rotations for ranges from $-7$ to $7$ degrees, and random slight cropping (Figure \ref{fig:additional-tta}). This captures naturalistic disturbances during real image capture.
\item \textbf{Random Photometric Augmentations:} This strategy simulates lighting changes, exposure, and color shifts. With adding brightness jitter, contrast jitter, color saturation jitter, and mild blur, the image embeddings are more robust to lighting and sensor variation.
\item \textbf{Multi-Crop Extraction:} Here, we use deterministic crops to capture different spatial regions, ensuring CLIP sees all quadrants and aiding with off-center target objects (Figure \ref{fig:multi-crop}).
\item \textbf{Grid Patches:} This technique acts similarly to the attention mechanism by splitting the image into 9 patches, capturing fine-local texture and allowing CLIP to see details that could be missed in large crops (Figure \ref{fig:grid-patch}).
\item \textbf{Mid-Quadrant Crops:} These image variations capture off-axis regions if target objects are not centered or in the corners (Figure \ref{fig:mid-quadrant-crop}).
\end{itemize}

Each candidate image $x_j$ (where $j\in \{1,2,..,10\}$) is transformed into a family of $m$ stochastic augmentations ${x_j}_1,{x_j}_2,...,{x_j}_m$. Each of these views is encoded independently by CLIP, producing $h_{{x_j}_1},h_{{x_j}_2},...,h_{{x_j}_m}$. Then, for each image, we take the mean of the embeddings across all augmentations $\overline{h_{x_j}}=\frac{1}{m}\sum_{i=1}^mh_{{x_j}_i}$, apply temperature($\tau$) scaling to the aggregated image embedding $\overline{h_{x_j}} * \frac{1}{\tau}$, and L2-normalize it to generate the final enriched image embedding $\tilde{h_{x_j}}$.

The result is a stable, view-averaged image embedding that is more semantically complete and less sensitive to occlusion and background noise. This improves the robustness and visual discrimination of our model. 

\subsubsection{Cosine Similarity}

\begin{figure}[ht]
    \begin{center}
        \includegraphics[width=1\linewidth]{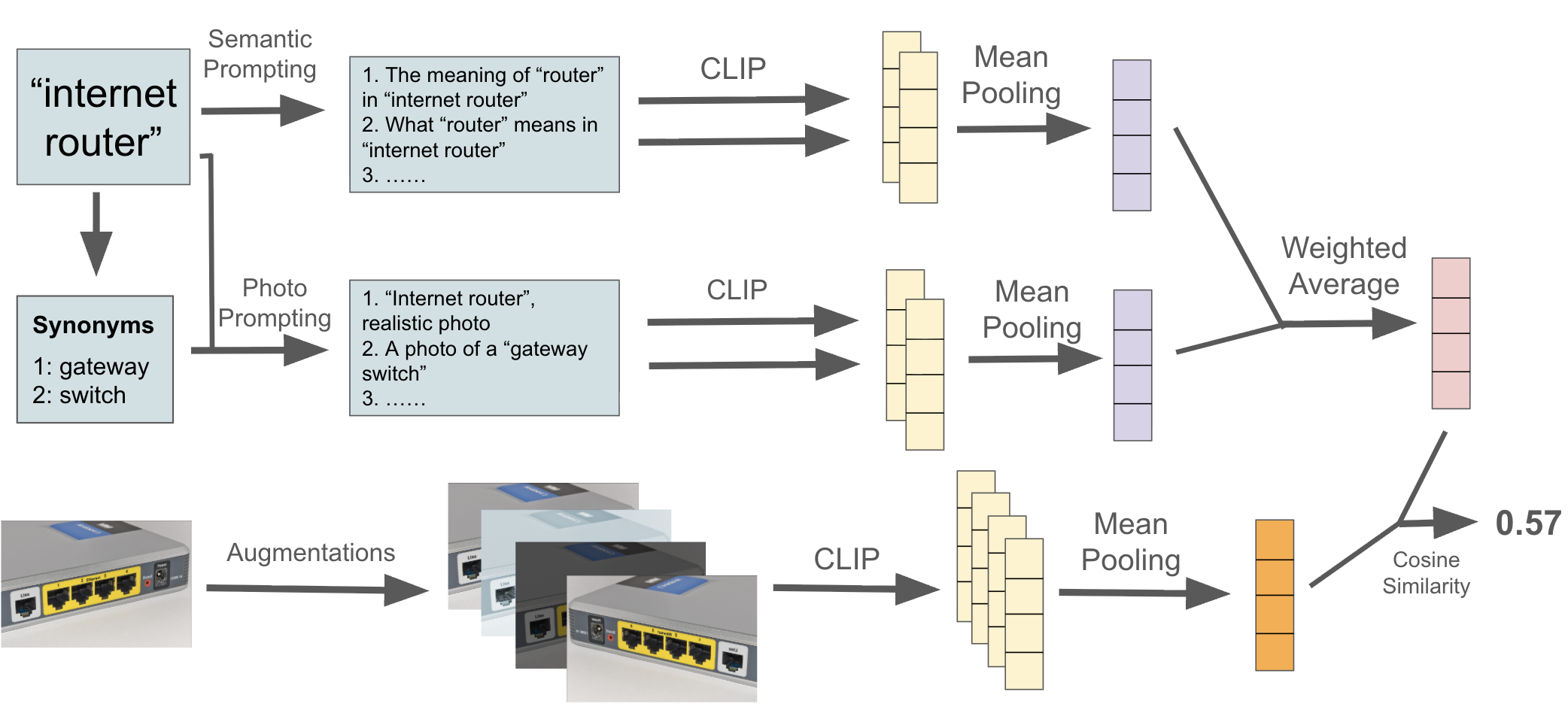}
        \caption{The final model. The target-sentence pair is expanded with a dual-channel prompt which consists of semantic and photo prompts. WordNet synonyms of both context and target words are also included for additional photo-prompts. All prompts are encoded by CLIP, averaged within each channel, fused with learned weights, and L2-normalized to produce robust textual embeddings. On the image side, each candidate image is transformed into a large set of stochastic multi-view augmentations. Each view is also encoded with CLIP, normalized, and the mean image embedding is computed. Cosine similarity between the final text embedding and each image embedding yields a score for ranking the images.}\label{fig:FinalModel}
    \end{center}
\end{figure}

In the final retrieval step, the enriched text embedding is compared to each of the enriched image embeddings using cosine similarity:
\begin{equation}
    x_*=\text{argmax}_{j\in \{1,2,..,10\}}\frac{\tilde{h_{t}}\cdot \tilde{h_{x_j}}}{\|\tilde{h_{d_*}}\|_2 \|\tilde{h_{x_j}}\|_2}
\label{eq:final_model}
\end{equation}

The top-ranked image $x_*$ is selected as the prediction for the target word’s sense in context. This model is visualized in Figure \ref{fig:FinalModel}.



\section{Additional Experiments}

\subsection{WordNet Definitions}
\label{wordnet_definitions}

To further enrich the textual embeddings $h_t$, we hypothesized that the model could use information from the WordNet definition that best defines the original target word. Therefore, following the approach in \cite{zhang2023srcb}, we queried WordNet to obtain a set of short definitions ${d_1,...,d_n}$ for the target word $t$. Each definition was embedded with CLIP to produce $h_{d_i}$, and the definitional embedding $h_{d_*}$ with the highest cosine similarity to the contextual embedding $h_t$ was selected as a lexical anchor for the target sense:

\begin{equation}
    d_*=\text{argmax}_{i\in \{1,2,..,k\}}\frac{h_t\cdot h_{d_i}}{\|h_t\|_2 \|h_{d_i}\|_2}
\label{eq:best_definition}
\end{equation}
Then, to balance contextual and definitional information, the model used the weighted average
\begin{equation}
h_{d,t}=\alpha h_{d_*}+(1-\alpha)h_t
\label{eq:definition_average}
\end{equation}
where $\alpha \in [0,1]$ controls the relative influence of WordNet versus the original contextual embedding. $h_{d,t}$ is subsequently normalized to remain compatible with cosine similarity.

Similar to Equation \ref{original_equation}, we choose the image that has the image embedding with the highest cosine similarity to the weighted definition embedding.

\begin{equation}
    x_*=\text{argmax}_{j\in \{1,2,..,10\}}\frac{h_{d,t}\cdot h_{x_j}}{\|h_{d,t}\|_2 \|h_{x_j}\|_2}
\label{eq:clip_with_wordnet}
\end{equation}



\subsubsection{Synonyms}
\label{synonom_definitions}

We also increased the number of candidate definitions by including the definitions of the top two synonyms of the target word. We theorized that this would improve accuracy in cases where richer definitional embeddings are present in the synonyms.

\subsection{BERT + BLIP Model}
\label{bert_blip_model}

\begin{figure}[H]
    \begin{center}
        \includegraphics[width=1\linewidth]{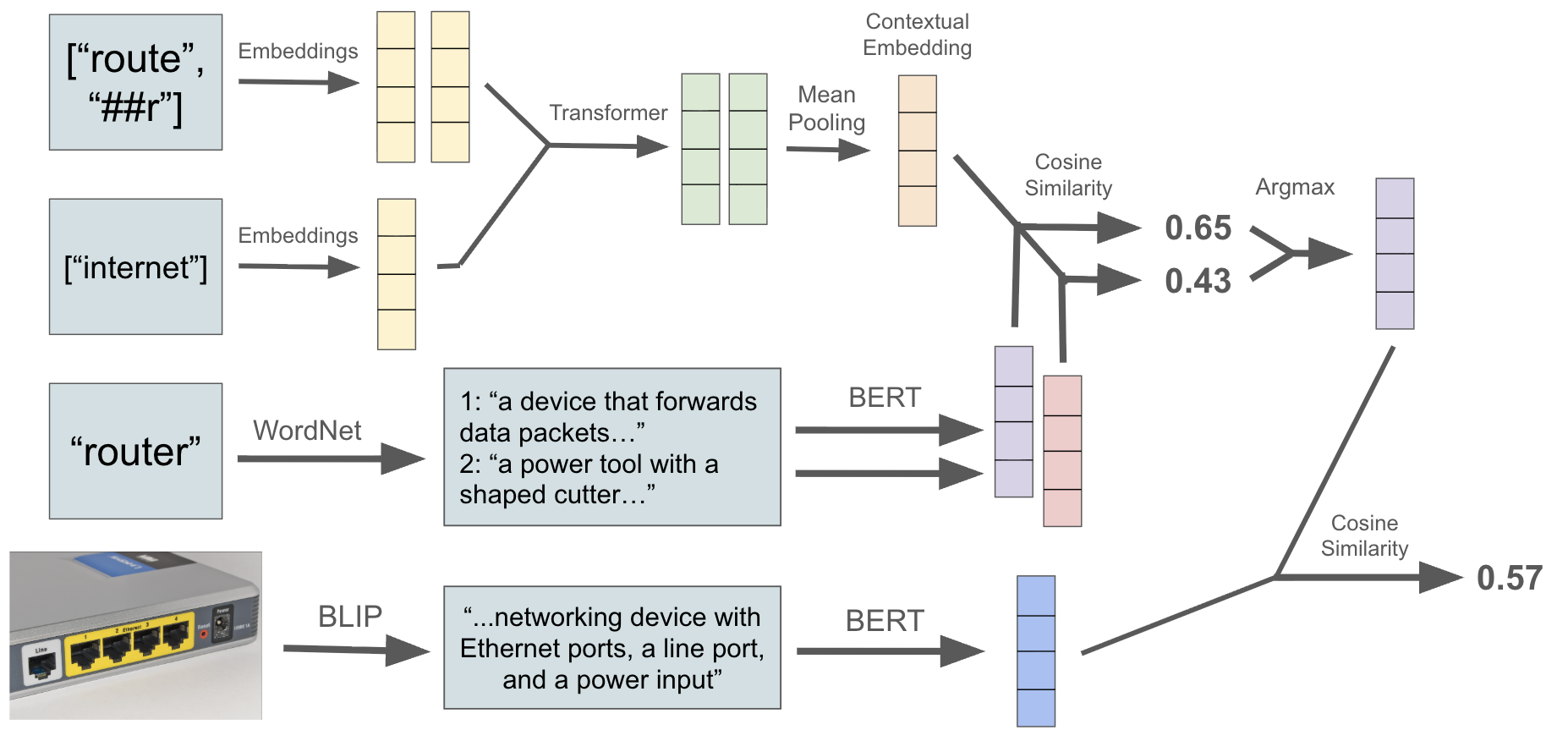}
        \caption{The BERT + BLIP model. As illustrated in Figure \ref{fig:ContextualEmbedding}, a contextual embedding of the target word is generated (orange). Similarly, embeddings for each definition of the word are produced (purple and red). The definitional embedding with the highest cosine similarity to the contextual embedding is stored (purple). Then, for the image, BLIP is used to generate a caption which is then passed into BERT for a textual embedding (blue). Finally, this embedding is compared to the chosen definitional embedding via cosine similarity.}\label{fig:OriginalModel}
    \end{center}
\end{figure}

To demonstrate the limitations of separate vision and language models, we constructed a baseline pipeline. Instead of using the VLM CLIP for the definitional embeddings, we employ the LLM BERT. Then, for the images, since BERT expects textual input, we first need to map the visual domain into the textual domain. Thus, the images $x_1,...,x_{10}$ were passed into the vision language model BLIP \cite{li2022blipbootstrappinglanguageimagepretraining} to generate textual captions $c_1,...,c_{10}$. Then, following the process explained in \ref{Contextual_Embeddings}, we used BERT to generate embeddings for each caption $h_{c_1},...,h_{c_{10}}$. Finally, we chose the image with the caption embedding that has the highest cosine similarity to the definition embedding (from Equation \ref{eq:best_definition}). In other words:
\begin{equation}
    x_*=\text{argmax}_{j\in \{1,2,..,10\}}\frac{h_{d_*}\cdot h_{c_j}}{\|h_{d_*}\|_2 \|h_{c_j}\|_2}
\label{eq:blip+bert}
\end{equation}
An illustration of this baseline model is shown in Figure \ref{fig:OriginalModel}.

\subsection{Translations}
\label{model_translation}
Another modification we explored was multilingual translation-based disambiguation. The hypothesis was that different languages often lexicalize ambiguities differently. For example, the English word ``bank" has two distinct Spanish translations: ``banco" (financial institution) versus ``orilla" (riverbank), forcing translation models to resolve ambiguity through word choice. 

To leverage this cross-linguistic disambiguation signal, we translated the semantic and photo prompts into Spanish, French, and German using MarianMT models from Helsinki-NLP. The CLIP text embeddings from both English and translated prompts were then averaged and normalized to create a multilingual representation.


\section{Results}
\label{results}

\subsection{The Dataset}

As explained in section \ref{dataset}, we test our model on the SemEval-2023 dataset. It is split into three sets. The trial set is for debugging and only has 16 sentences. The train set is used for hyperparameter selection and model development with 12,869 examples. The test set is used for model evaluation and has 463 sentences. Unless otherwise specified, the results for our model are trained on the train set and evaluated on the test set. 

\subsection{Evaluation Metrics}

To evaluate the model’s performance, we utilize two well-established metrics. Let $N$ be the number of samples in the dataset and $\mathds{1}_1$ be the indicator function, which returns $1$ if the expression inside is $1$ and $0$ otherwise. Also, for the $i$th target word, define $r_i$ to be the predicted rank (out of 10) of the correct image. For example, if $r_i=1$, the selected image was correct. But, if $r_i=3$ it means that $2$ other candidate images had higher predicted relevance scores.

The hit rate measures the overall accuracy of the predictions:
\[
\text{Hit Rate}=\frac{1}{N}\sum_{i=1}^N \mathds{1}_1[r_i]
\]

The Mean reciprocal rank (MRR) measures the average inverse rank, rewarding models that rank the correct image higher:
\[
\text{MRR}=\frac{1}{N}\sum_{i=1}^N\frac{1}{r_i}
\]

\subsection{Final Model}


With all text-side enrichments and image augmentations combined into a unified CLIP framework (Section \ref{proposed_architecture}), we perform Bayesian hyperparameter tuning using an 80/20 internal train–validation split on the training data. We employ the Python library optuna to search over the dual-channel weights ($\beta_p,\beta_s$) and temperature scaling parameter ($\tau$). The best-performing hyperparameter configuration is then evaluated on the held-out test set to assess generalization. 

The optimal hyperparameters and the final results are illustrated in Table  \ref{tab:bayes_params} and \ref{tab:final_performance} respectively.

\begin{table}[h]
    \centering
    \begin{tabular}{lccc}
    \hline
    \textbf{Model} &
    $\mathbf{\beta_p}$ &
    $\mathbf{\beta_s}$ &
    $\mathbf{\tau}$ \\
    \hline
    CLIP-ViT-B/32 &
    0.7685 & 0.6152 & 0.6335 \\
    
    CLIP-ViT-B/16 &
    0.8235 & 0.3079 & 0.8342 \\
    
    CLIP-ViT-B/32 (LAION) &
    0.7960 & 0.1042 & 0.8919 \\
    \hline
    \end{tabular}
    \caption{Best Hyperparameters from Bayesian Optimization}
    \label{tab:bayes_params}
\end{table}




\begin{table}[h]
\centering
\label{tab:final_performance}
\begin{tabular}{lcc}
\hline
\textbf{Model} &
\textbf{Final MRR} &
\textbf{Final Hit Rate} \\
\hline
CLIP-ViT-B/32 &
0.7392 & 0.5940 \\

CLIP-ViT-B/16 &
0.7522 & 0.6177 \\

\textbf{CLIP-ViT-B/32 (LAION)} &
\textbf{0.7590} & \textbf{0.6220} \\
\hline
\end{tabular}
\caption{Final Evaluation Performance Metrics on Test Set}
\label{tab:final_performance}
\end{table}

Enriching the embeddings with dual-channel prompting and image augmentations increased the MRR from 0.7227 to 0.7590 and the Hit Rate from 0.5810 to 0.6220.

\subsection{Baselines}

\subsubsection{Vanilla CLIP-based Model}
\label{CLIP_results}
To understand how well the model performs before incorporating prompting, WordNet, image augmentations, we tested the vanilla CLIP-based model (Section \ref{vanilla_model}) on the train and test sets. The MRR and Hit Rate metrics are displayed in Table \ref{tab:clip_comparison}.

\begin{table}[H]
    \centering
    \begin{tabular}{lcc}
    \hline
    \textbf{Approach} & \textbf{MRR} & \textbf{Hit Rate} \\
    \hline
    Train Set & 0.8268 & 0.7316 \\
    Test Set & 0.7227 & 0.5810 \\
    \hline \\
    \end{tabular}
\caption{The MRR and Hit Rate of the vanilla CLIP model on the English train and test sets.}
\label{tab:clip_comparison}
\end{table}

The vanilla CLIP-based model performs well but fails to meet state-of-the-art test results.

\subsubsection{BERT + BLIP Model}
\label{BERT_BLIP_Results}

To investigate the importance of using a vision language model such as CLIP, which maps textual and visual input to the same embedding space, we tested the BERT + BLIP model explained in Section \ref{bert_blip_model}. Initial testing on a small subset of the SemEval-2023 dataset yielded a hit rate of 0\%. Additionally the cosine similarity scores were low and the predictions appeared essentially random. 

This discrepancy arose from differences in the embedding spaces of the models used. Since BLIP is not specifically designed to project images and text into a shared embedding space, the generated captions—though semantically accurate—produced vector representations that were misaligned with those of the target words. As a result, the computed cosine similarities lacked meaningful interpretability, leading to inconsistent or nonsensical outcomes. This underscores the necessity of using CLIP to ensure meaningful cross-modal similarity comparisons.

\subsection{Ablation Studies}

We analyze the trade-offs between latency cost and retrieval quality across isolated and combined configurations in order to understand how semantic guidance and visual augmentations contribute to performance and scalability in multimodal disambiguation. The configurations we analyze include, prompting only, visual augmentations only, their combination, and their combination with a 80/20 validation split on the training set.

\begin{table}[H]
    \centering
    \label{tab:prompting_only}
    \begin{tabular}{l c}
    \hline
    \textbf{Component} & \textbf{Value} \\
    \hline
    Prompting & Enabled \\
    Augmentations & Disabled \\
    Views per image & 1 \\
    Candidates per query & 10 \\
    \hline
    Text embedding latency (mean) & 39.43 ms \\
    Image embedding latency (mean, per image) & 5.28 ms \\
    Estimated image embedding per query & 52.81 ms \\
    End-to-end ranking latency (mean, per query) & 62.95 ms \\
    \hline
    MRR & 0.7506 \\
    Hit@1 & 0.6250 \\
    \hline
    \end{tabular}
    \caption{Computational cost and retrieval performance for the prompting-only configuration. Reported latencies are mean values.}
    \label{res:prompting_only}
\end{table}

The prompting-only configuration incorporates textual guidance without any visual augmentations. On this iteration of testing we set $\tau=0.7$, $\beta_p=0.6$, and $\beta_s=0.4$. As shown in Table \ref{res:prompting_only}, with a single view per image, the setup achieves a mean end-to-end latency of 62.95 ms per query. This, as will be seen later, is over an order of magnitude faster than the augmentation-based method. Even with its low computational cost, prompting-only retrieval achieves strong performance (MRR: 0.751, Hit@1: 0.625), indicating that semantic cues provide significant disambiguation power. This result demonstrates that incorporating prompt-based semantics is an effective strategy. 

\begin{table}[H]
    \centering
    \label{tab:augmentation_cost}
    \begin{tabular}{l c}
    \hline
    \textbf{Component} & \textbf{Value} \\
    \hline
    Prompting & Disabled \\
    Augmentations & Enabled \\
    Views per image & 28 \\
    Candidates per query & 10 \\
    \hline
    Text embedding latency (mean) & 23.68 ms \\
    Image embedding latency (mean, per image) & 59.65 ms \\
    Estimated image embedding per query & 596.47 ms \\
    End-to-end ranking latency (mean, per query) & 709.77 ms \\
    \hline
    MRR & 0.7225 \\
    Hit@1 & 0.5745 \\
    \hline
    \end{tabular}
    \caption{Computational cost and retrieval performance under the augmentation-only configuration. Reported latencies are mean values.}
    \label{fig_augmentation_only}
\end{table}

The augmentations-only approach isolates the contribution of visual diversity by embedding different  views of each visual. Without any semantic prompting, this setup uses 28 augmented views per candidate image, incurring substantial mean end-to-end latency cost of approximately 710 ms per query driven mostly by image embeddings. However, as shown in Figure \ref{fig_augmentation_only}, this high expense does not contribute to a much higher performance from the base CLIP model, demonstrating that visual augmentation alone can only somewhat mitigate ambiguity in visually complex scenarios. When compared to prompting-based approaches, however, this result indicates visual diversity is a less efficient source or disambiguation compared to prompting.

\begin{table}[H]
    \centering
    \label{tab:prompting_augmentations}
    \begin{tabular}{l c}
    \hline
    \textbf{Component} & \textbf{Value} \\
    \hline
    Prompting & Enabled \\
    Augmentations & Enabled \\
    Views per image & 28 \\
    Candidates per query & 10 \\
    \hline
    Text embedding latency (mean) & 13.25 ms \\
    Image embedding latency (mean, per image) & 59.44 ms \\
    Estimated image embedding per query & 594.41 ms \\
    End-to-end ranking latency (mean, per query) & 714.16 ms \\
    \hline
    MRR & 0.7315 \\
    Hit@1 & 0.5853 \\
    \hline
    \end{tabular}
    \caption{Computational cost and retrieval performance for the combined prompting and augmentation configuration. Reported latencies are mean values.}
    \label{table_both}
\end{table}

On this iteration of testing we set $\tau=0.7$, $\beta_p=0.6$, and $\beta_s=0.4$. As shown in Table \ref{table_both}, combining prompting with multi-view image augmentation unsurprisingly increases computational costs while under performing in both retrieval performance and end-to-end latency compared to the prompting-only method. This suggests diminishing returns from aggressive visual augmentation when semantic guidance is present, highlighting an important balance between visual diversity and inference efficiency. 

\begin{table}[H]
    \centering
    \label{tab:prompting_augmentations_split}
    \begin{tabular}{l c}
    \hline
    \textbf{Component} & \textbf{Value} \\
    \hline
    Prompting & Enabled \\
    Augmentations & Enabled \\
    Views per image & 28 \\
    Candidates per query & 10 \\
    \hline
    Text embedding latency (mean) & 17.93 ms \\
    Image embedding latency (mean, per image) & 59.35 ms \\
    Estimated image embedding per query & 593.47 ms \\
    End-to-end ranking latency (mean, per query) & 714.38 ms \\
    \hline
    Validation MRR & 0.8511 \\
    Validation Hit@1 & 0.7708 \\
    Test MRR & 0.7272 \\
    Test Hit@1 & 0.5767 \\
    \hline
    \end{tabular}
    \caption{Computational cost (mean latencies) and retrieval performance for prompting with augmentations using a validation/test split.}
    \label{table_both_validation}
\end{table}

To ensure our model generalizes and avoid over-fitting, we evaluate the prompting-plus-augmentation configuration using a validation/test split. On this iteration of testing we set $\tau=0.7$, $\beta_p=0.6$, and $\beta_s=0.4$. As shown in Table \ref{table_both_validation}, the validation set achieves a high performance (MRR: 0.851, Hit@1: 0.771), indicating a strong alignment under tuned conditions. However, performance drops significantly on the test set (MRR: 0.727, Hit@1: 0.577), showing similar results as the augmentation-only baseline. This gap demonstrates that while prompting and augmentation can be optimized to fit validation data, their combined benefits do not fully transfer over to unseen examples, suggesting overfitting or dataset-specific prompt sensitivity. 

From a high-level, these results give more insight on the roles of semantic prompting and visual augmentation in multimodal retrieval. Overall, prompting provides a strong and efficient signal, while image augmentations provide more visual coverage at a higher computational cost. Under realistic test conditions, the combination of both prompting and image augmentations provide limited gains, emphasizing that retrieval quality is contained by generalization and fusion strategy as well. 

\subsection{Additional Experiments}

\subsubsection{WordNet}

Table \ref{tab:wordnet_results} displays the results when WordNet definitions are integrated into the model (Section \ref{wordnet_definitions}). 

\begin{table}[H]
    \centering
    \begin{tabular}{lcc}
    \hline
    \textbf{Approach} & \textbf{MRR} & \textbf{Hit Rate} \\
    \hline
    Baseline (0\% WordNet) & 0.7227 & 0.5810 \\
    5\% WordNet & 0.7202 & 0.5745 \\
    10\% WordNet & 0.7284 & 0.5896 \\
    \textbf{15\% WordNet} & \textbf{0.7296} & \textbf{0.5896} \\
    20\% WordNet & 0.7283 & 0.5874 \\
    25\% WordNet & 0.7280 & 0.5896 \\
    35\% WordNet & 0.7247 & 0.5875 \\
    50\% WordNet & 0.7049 & 0.5659 \\
    75\% WordNet & 0.6641 & 0.5184\\
    100\% WordNet & 0.6068 & 0.4622\\
    \hline \\
    \end{tabular}
\caption{The MRR and Hit Rate on the test set of the model that uses definitions from WordNet. The model performs the best when the embedding is 15\% of the WordNet definitional embedding and 85\% of the original contextual embedding ($\alpha=0.15$).}
\label{tab:wordnet_results}
\end{table}

If we replace the contextual embedding entirely with the definitional embedding, the model doesn't perform well. This likely occurs due to inaccuracies in choosing the best definition. However, when we define our embedding to be a weighted average of the contextual embedding and definitional embedding (Equation \ref{eq:definition_average}), results improve. This is because we maintain most of the information from the original sentence while also refining its embedding with more information from a detailed definition.



In Table \ref{tab:synonoms} the results of the WordNet model that also uses definitions of the synonyms for candidate definitional embeddings (Section \ref{synonom_definitions}) are displayed. The MRR and hit rate decrease slightly, likely due to the fact that the definitions of synonyms are not as well representative of the intended meaning of the target word.

\begin{table}[H]
    \centering
    \begin{tabular}{lcc}
    \hline
    \textbf{Approach} & \textbf{MRR} & \textbf{Hit Rate} \\
    \hline
    15\% WordNet & 0.7254 & 0.5832 \\
    \textbf{25\% WordNet} & \textbf{0.7255} & \textbf{0.5832} \\
    50\% WordNet & 0.7023 & 0.5616 \\
    100\% WordNet & 0.6025 & 0.4557\\
    \hline \\
    \end{tabular}
\caption{The MRR and Hit Rate on the test set of the model that uses definitions from WordNet of both the target word and two synonyms.}
\label{tab:synonoms}
\end{table}

Overall, the gains from incorporating WordNet definitions were not statistically significant enough to be included in our final model.

\subsubsection{Translation Results}

We tested the implementation of translation into the model (Section~\ref{model_translation}) across single-language configurations— ES (English), FR (French), and DE (German)—two-language combinations (ES+FR), and a three-language ensemble (ES+FR+DE).


\label{translation_results}

\begin{table}[H]
    \centering
    \begin{tabular}{lcc}
    \hline
    \textbf{Sentence Translation Experiment} & \textbf{MRR} & \textbf{Hit Rate} \\
    \hline
    \textbf{Baseline (English only)} & \textbf{0.7217} & \textbf{0.5788} \\
    Spanish & 0.6989 & 0.5508 \\
    French & 0.6958 & 0.5443 \\
    German  & 0.6729 & 0.5162 \\
    Multi (ES+FR+DE) & 0.6756 & 0.5313 \\
    Multi (ES+FR) & 0.6827	& 0.5292 \\
\hline \\
    \end{tabular}
\caption{Impact of multilingual sentence translation on visual word sense disambiguation performance on the test set. English-only baseline achieves the highest MRR (0.7217) and Hit Rate (0.5788).}
\label{tab:translation_comparison}
\end{table}

Tables \ref{tab:translation_comparison} and \ref{tab:translation_all_comparison} demonstrate that multilingual translation (Section \ref{model_translation}) consistently degrades performance across all metrics. The English-only baseline achieves the highest MRR (72.17\%) and Hit Rate (57.45\%), outperforming all translation-based approaches. Performance degradation follows a clear pattern: single-language translations (Spanish: 69.74\% MRR, French: 69.87\% MRR, German: 68.48\% MRR) already show 2-4\% decreases, while multi-language combinations exhibit further losses, with the three-language ensemble (ES+FR+DE) achieving only 66.92\% MRR, a 5.3\% decrease from baseline. This progressive degradation suggests that averaging embeddings across languages introduces more noise than a disambiguating signal. 

\begin{table}[H]
    \centering
    \begin{tabular}{lcc}
    \hline
    \textbf{Prompting with Translations Experiment} & \textbf{MRR} & \textbf{Hit Rate} \\
    \hline
    \textbf{Baseline (All Prompts + Rerank) } & \textbf{0.7217} & \textbf{0.5745} \\
    Baseline (All Prompts, No Rerank) & 0.7143 & 0.5680 \\
    Spanish (All Prompts + Rerank) & 0.6974 & 0.5443 \\
    French (All Prompts + Rerank)   & 0.6987 & 0.5464 \\
    German (All Prompts + Rerank)   & 0.6848 & 0.5248 \\
    Multi ES+FR (All Prompts + Rerank)  & 0.6805	& 0.5270 \\
    Multi ES+DE (All Prompts + Rerank)  & 0.6785	& 0.5205 \\
    Multi DE+FR (All Prompts + Rerank)  & 0.6775	& 0.5184 \\
    Multi ES+FR+DE (All Prompts + Rerank)  & 0.6692	& 0.5097 \\
    Multi DE+FR (No Rerank)  & 0.6709	& 0.5162 \\
    Multi ES+FR+DE (No Rerank)  & 0.6530	& 0.4946 \\
\hline \\
    \end{tabular}
\caption{Prompting with multilingual translation results on the test set. Baseline English prompting with re-ranking outperforms all multilingual configurations.}
\label{tab:translation_all_comparison}
\end{table}

Several factors contribute to this negative result. First, due to MarianMT translation errors, ambiguous words may be mistranslated or lose critical contextual nuances. Secondly, the embedding averaging process creates a ``diluted" representation that lacks the specificity needed for fine-grained disambiguation, effectively smoothing away the semantic distinctions necessary to identify the correct image. The marginal improvements from re-ranking (approximately 1-1.6\% across configurations) prove insufficient to overcome the fundamental translation-induced performance loss. This confirms that for CLIP-based visual word sense disambiguation, maintaining high-quality monolingual English representations yields superior results to multilingual ensemble approaches.




\section{Conclusion}


This project investigates how large language models handle lexical ambiguity where a single word can have multiple meanings. We focus on Visual Word Sense Disambiguation (VWSD), where the goal is to choose which image best represents a target word in its given textual context. Our proposed model leverages OpenAI's CLIP model to embed text and images into a shared multi-modal space, refines those embeddings with a dual-channel prompting ensemble and image augmentations, and compares them via cosine similarity.

Our ablation studies reveal that model performance is highly sensitive to how semantic and visual information is introduced and combined. Prompt-based semantic guidance proved to the the most effective and computationally efficient signal, achieving a relatively stronger retrieval performance with minimal inference cost. In contrast, aggressive image augmentation introduced substantial computational overhead and yielded significantly lower returns when used without careful fusion. An informed balance of these strategies is needed to truly increase performance. 

Further ablations that integrate external knowledge added additional challenges. Our ablation studies reveal that multilingual translations and excessive WordNet integration introduced significant noise. Together, these findings highlight that adding more information does not guarantee better disambiguation unless that information perceptively aligned with the task.


While our final model achieved an MRR of 0.7590 and a Hit Rate of 0.6220 on the test set, a notable gap remains between our results and the top-performing SemEval-2023 systems, which reported Hit@1 scores to be approximately 84-85\%. This gap arises from several factors including the use of larger and more specialized models, learned fusion mechanisms that are beyond simple cosine similarity, and more complex retrieval or re-ranking strategies. 

Our approach prioritizes interpretability, controlled ablation, and modularity, while being aware of performance cost due to limited resource accessibility, providing clear insight into how individual components contribute to disambiguation. This creates a foundation for future work to close the state-of-the-art system gap while maintaining sustainable inference costs.


\bibliographystyle{plainnat}
\bibliography{ref}

\appendices

\section{Image Augmentations}

\label{app:augmentations}

The images below are examples of some of the augmentations our model generates for the image of an internet router:

\begin{figure}[H]
\centering
\begin{tabular}{cc}
    \includegraphics[width=0.3\linewidth]{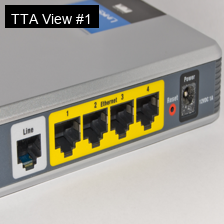} &
    \includegraphics[width=0.3\linewidth]{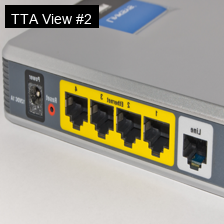} \\
    \small Original Image & \small Horizontal Flip \\[1em]
\end{tabular}
\caption{Examples of the Test-Time Augmentation Strategy.}
\label{fig:deterministic-tta}
\end{figure}

\end{document}